\documentclass[letterpaper, 10 pt, conference]{ieeeconf}  

\IEEEoverridecommandlockouts                              

\title{\LARGE \bf
An Efficiently Solvable Quadratic Program for Stabilizing\\ Dynamic Locomotion 
}

\author{Scott Kuindersma, Frank Permenter, and Russ Tedrake
\thanks{This work was supported by AFRL contract FA8750-12-1-0321 and NSF contract ERC-1028725, IIS-1161909, and IIS-0746194.}%
\thanks{The authors are with the Computer Science and Artificial Intelligence Laboratory at the Massachusetts Institute of Technology, Cambridge, MA, USA.
        {\tt\small \{scottk,fpermenter,russt\}@csail.mit.edu}}%
}

\usepackage{bm,amsfonts,subfigure,graphicx,algorithm2e,amsmath,pgfplots,url}

\newcommand{\stack}[2]{\underset{#2}{\operatorname{#1}} \;}
\newcommand{\vecg}[1]{\bm{#1}}
\renewcommand{\vec}[1]{\mathbf{#1}}

\begin{document}

\maketitle
\thispagestyle{empty}
\pagestyle{empty}

\begin{abstract}
We describe a whole-body dynamic walking controller implemented as a convex quadratic program. The controller solves an optimal control problem using an approximate value function derived from a simple walking model while respecting the dynamic, input, and contact constraints of the full robot dynamics. By exploiting sparsity and temporal structure in the optimization with a custom active-set algorithm, we surpass the performance of the best available off-the-shelf solvers and achieve 1kHz control rates for a 34-DOF humanoid. We describe applications to balancing and walking tasks using the simulated Atlas robot in the DARPA Virtual Robotics Challenge. 
\end{abstract}

\section{Introduction}

Achieving dynamically-stable locomotion in complex legged systems is a problem at the heart of modern robotics research. For humanoid systems in particular, nonlinear, underactuated, and high-dimensional dynamics conspire to make the control problem challenging. Optimization-based techniques must simultaneously reason about the dynamics, actuation limits, and contact constraints of the walking system. Model predictive control (MPC) is a popular approach to performing this type of constrained optimization iteratively over fixed horizons, but its computational complexity has hindered applications to high-dimensional systems. Furthermore, the hybrid dynamics of walking robots makes multi-step optimization difficult~\cite{PosaTedrake2012}. Successful examples of using MPC for humanoid control have therefore relied upon the use of low-dimensional linear models \cite{StephensAtkeson2010, Dimitrov_etal2011} or relaxation of constraints to permit smooth optimization through discontinuous dynamics \cite{Tassa_etal2012}. 

Several researchers have recently explored using quadratic programs (QPs) to control bipedal systems by exploiting the fact that the \emph{instantaneous} dynamics and contact constraints can be expressed linearly (effectively solving a horizon-1 MPC problem) \cite{Abe07, Collette_etal07, Macchietto_etal2009, Ames2012, Herzog_etal2013, Kudoh_etal2002, Saab_etal2013, Koolen_etal2013}. 
A key observation about these approaches in the context of balancing and locomotion tasks is that, during typical operation, the set of active inequality constraints changes very infrequently between consecutive control steps. We give a problem formulation and solution technique that explicitly take advantage of this observation. 

We describe a QP that exploits optimal control solutions for a simple unconstrained model of the walking system. Using time-varying LQR design, we compute the optimal cost-to-go for the simple model and use it as part of the objective function in a constrained optimization to compute inputs for the full robot. We describe the approach concretely in terms of a simulated bipedal system and zero-moment point (ZMP) dynamics. In addition to providing a principled and reliable way to stabilize walking trajectories, we show the resulting QP cost function contains low-dimensional structure that can be exploited to reduce solution time. 

To achieve real-time control rates, we designed a custom active-set solver that exploits consistency between subsequent solutions and outperforms the best available off-the-shelf solvers such as CVXGEN and Gurobi by a factor of 5 or more. Our analysis of solver performance during typical walking experiments suggests that the active set remains constant between consecutive control steps approximately 97\% of the time, requiring only a \emph{single linear system solve per step}. In our tests, we were able to achieve average control rates of 1kHz for a 34-DOF humanoid. We briefly summarize extensive simulation testing done with the Atlas robot as part of the DARPA Virtual Robotics Challenge.


\section{LQR Design for ZMP Dynamics}

The planar center of mass (COM) and ZMP dynamics of a fully actuated rigid body system can be written in state space form as
\begin{eqnarray}
\dot{\vec x} &=& \vec A \vec x + \vec B \vec u \label{com_dynamics} \nonumber\\
 &=& \left [ \begin{array}{cc}
0 & \vec I \\
0 & 0
\end{array} \right ] \vec x + \left [ \begin{array}{c}
0 \\
\vec I
\end{array} \right ] \vec u \\
\vec y &=&  \vec C \vec x - b(\vec x, \dot{\vec x})\vec u\nonumber\\
 &=& \left [ \begin{array}{cc}
\vec I & 0 
\end{array} \right ] \vec x + \frac{z_{\rm com}}{\ddot{z}_{\rm com} + g} \vec I \vec u, \label{eq:zmp} 
\end{eqnarray}
where $\vec x = [x_{\rm com}, y_{\rm com},\dot{x}_{\rm com}, \dot{y}_{\rm com}]^T$, $\vec u = [\ddot{x}_{\rm com}, \ddot{y}_{\rm com}]^T$, $\vec y = [x_{\rm zmp}, y_{\rm zmp}]^T$, $g$ is a constant gravitational acceleration, and $z_{\rm com}$ is the COM height. 
The ZMP is a well-studied quantity in the bipedal walking literature that defines the point on the ground plane at which the moment produced by inertial and gravitational forces is parallel to the surface normal (i.e., the robot is not tipping)~\cite{SardainZMP}. Since dynamic balance is achieved when the contact forces directly oppose the gravitational and inertial forces, maintaining the ZMP within the contact support polygon can be an effective strategy for maintaining dynamic stability in legged locomotion. 

Given desired ZMP trajectory, $\vec y^d(t)$, we would like to compute an optimal tracking controller that takes into account the time- and state-varying constraints on $\vec u$ imposed by the dynamics, input limits, and contacts of the full walking system. Due to the prohibitive computational requirements of solving nonlinearly constrained optimal control problems of this scale, we instead solve an unconstrained time-varying LQR problem to compute the optimal cost-to-go, $J^*$, which provides a control-Lyapunov function (CLF) for the ZMP dynamics. On each iteration, we select the control inputs to descend this ZMP CLF while reasoning about the instantaneous constraints of the full system. 

We begin by specifying a cost functional of the form 
\begin{eqnarray}
J = \bar{\vec y}(t_f)^T \vec Q_f \bar{\vec y}(t_f) + \int_0^{t_f} \bar{\vec y}(t)^T \vec Q \bar{\vec y}(t) dt, \label{tvlqr_cost}
\end{eqnarray}
where the coordinates $\bar{\vec y}(t) = \vec y(t)-\vec y^d(t)$, $\vec Q \succ 0$, and $\vec Q_f \succ 0$. 
In practice the COM height, $z_{\rm com}$, is often assumed to be constant, making the ZMP dynamics (\ref{eq:zmp}) linear~\cite{Kajita_etal03}. More generally, if the COM height trajectory is constrained to be a known function of time, $(z_{\rm com}(t),\dot{z}_{\rm com}(t),\ddot{z}_{\rm com}(t))$, the ZMP dynamics are time-varying linear,
\begin{eqnarray}
\vec y(t) = \vec C(t) \vec x(t) + \vec D(t) \vec u(t), \label{eq:tvzmp}
\end{eqnarray}
and therefore amenable to TVLQR design without explicit linearization.
  
Solving the Riccati equation yields the optimal cost-to-go for the time-varying linear system, 
\begin{eqnarray*}
J^*(\bar{\vec x},t) = \bar{\vec x}^T \vec S(t) \bar{\vec x} + \vec s_1(t)^T \bar{\vec x} + s_0(t),
\end{eqnarray*}
and the linear optimal controller,
\begin{eqnarray}
\bar{\vec u}^* &=& -\vec K(t) \bar{\vec x} \nonumber\\
&=& \arg \min_{\bar{\vec u}} \bar{\vec y}(t)^T \vec Q \bar{\vec y}(t) + \frac{\partial J^*}{\partial \bar{\vec x}}\bigg|_{\bar{\vec x}} \dot{\bar{\vec x}}, \label{eqn:HJB}
\end{eqnarray}
where $\bar{\vec x}(t) = \vec x(t) - \vec x^d(t)$ and $\bar{\vec u}(t) = \vec u(t) - \vec u^d(t)$.
In general, achieving $\bar{\vec u}^*$ is not possible due to constraints imposed by the robot dynamics. For example, actuator saturations and contact friction properties can limit the possible magnitudes and directions of COM accelerations. Therefore, to compute control inputs we perform a constrained minimization using 
\begin{eqnarray}
V(\bar{\vec x},\bar{\vec u},t)= \bar{\vec y}(t)^T \vec Q \bar{\vec y}(t) + \frac{\partial J^*}{\partial \bar{\vec x}}\bigg|_{\bar{\vec x}} \dot{\bar{\vec x}} \label{eqn:hjbcost}
\end{eqnarray}
as a surrogate value function.

\section{QP Formulation}\label{sec:QP}

Given the stabilizing solution for the ZMP dynamics, we design a QP to solve for control inputs for the full robot dynamics that minimizes (\ref{eqn:hjbcost}) and a quadratic motion cost for walking subject to the instantaneous constraints. 

Consider the familiar rigid body dynamics,
\begin{eqnarray}
\vec H(\vec q) \ddot{\vec q} + \vec C(\vec q,\dot{\vec q}) = \vec B(\vec q,\dot{\vec q}) \vecg \tau + \vecg \Phi(\vec q)^T \vecg \lambda, \label{eqn:manip-dynamics}
\end{eqnarray}
where $\vec H(\vec q)$ is the system inertia matrix, $\vec C(\vec q,\dot{\vec q})$ captures the gravitational and Coriolis terms, $\vec B(\vec q,\dot{\vec q})$ is the control input map, and $\vecg \Phi(\vec q)^T$ transforms external forces, $\vecg \lambda$, into generalized forces. In our case, $\vecg \lambda = [\begin{array}{ccc} \vecg \lambda_1^T & \dots & \vecg \lambda_{N_c}^T \end{array} ]^T$ is a vector of ground-contact forces acting at $N_c$ contact points. The set of active contacts are determined by kinematic or force measurement classification at each control step.  

For floating-base systems such as humanoids, the dynamics can be partitioned into actuated and unactuated degrees of freedom~\cite{Herzog_etal2013},
\begin{eqnarray}
\vec H_f \ddot{\vec q} + \vec C_f &=&  \vecg \Phi_f^T \vecg \lambda \label{eqn:float_dyn} \\
\vec H_a \ddot{\vec q} + \vec C_a &=& \vec B_a \vecg \tau + \vecg \Phi_a^T \vecg \lambda \nonumber, 
\end{eqnarray}
where we have dropped the explicit dependence on $\vec q, \dot{\vec q}$ from our notation for conciseness. This separation permits the removal of $\vecg \tau$ as a decision variable by including \eqref{eqn:float_dyn} as a constraint expressing $\vecg \tau$ in terms of $\ddot{\vec q}$ and $\vecg \lambda$:
\begin{eqnarray*}
{\vecg \tau} = \vec B_a^{-1}\left[\vec H_a \ddot{\vec q} + \vec C_a - \vecg \Phi_a^T \vecg \lambda\right].
\end{eqnarray*}

We use a standard, conservative polyhedral approximation of the friction cone, $\hat{K}_j$, for each contact point, $\vec c_j$,  
\begin{eqnarray}
\hat{K}_j = \left \{ \sum_{i=1}^{N_d} \beta_{ij} \vec v_{ij}: \beta_{ij} \ge 0 \right \}.\label{k_approx}
\end{eqnarray}
The generating vectors, $\vec v_{ij}$, are computed as $\vec v_{ij}=\vec n_j + \mu_j \vec d_{ij}$, where $\vec n_j$ and $\vec d_{ij}$ are the contact-surface normal and $i^{\rm th}$ tangent vector for the $j^{\rm th}$ contact point, respectively,  $\mu_j$ is the Coulomb friction coefficient, and $N_d$ is the number of tangent vectors used in the approximation~\cite{PollardReitsma01}. 

Given the robot state, $\vec q, \dot{\vec q}$, at time $t$, we solve the following quadratic program:

\newtheorem{qp}{Quadratic Program}
\begin{qp}
\begin{eqnarray}
\min_{\ddot{\vec q},\vecg \beta, \vecg \lambda, \vecg \eta} V(\bar{\vec x},\bar{\vec u},t) + w_{\ddot{\vec q}}|| \ddot{\vec q}_{\rm des}-\ddot{\vec q} ||^2 + \varepsilon \sum_{ij} \beta^2_{ij} + ||\vecg \eta ||^2 \label{eqn:qpcost}
\end{eqnarray}
subject to
\begin{eqnarray}
\vec H_f \ddot{\vec q} + \vec C_f &=& \vecg \Phi_f^T \vecg \lambda \label{con:dynamics}\\
\vec J \ddot{\vec q} + \dot{\vec J} \dot{\vec q} &=& -\alpha \vec J \dot{\vec q} + \vecg \eta \label{con:accel} \\
\vec B_a^{-1}(\vec H_a \ddot{\vec q} + \vec C_a - \vecg \Phi_a^T \vecg \lambda) &\in& [\vecg \tau_{\rm min},\vecg \tau_{\rm max}] \label{con:inputs} \\
\forall_{j=\{1\dots N_c\}}~~ \vecg \lambda_j &=& \sum_{i=1}^{N_d} \beta_{ij} \vec v_{ij} \label{con:contact_force} \\
\forall_{i,j} \beta_{ij} &\ge& 0  \label{con:friction} \\
\vecg \eta &\in& [\vecg \eta_{\rm min},\vecg \eta_{\rm max}]. \label{con:slack}
\end{eqnarray} \label{qp:controller}
\end{qp}
The constraints (\ref{con:dynamics}) and (\ref{con:inputs}) ensure that the dynamics and input limits are respected, (\ref{con:accel}) is a no-slip constraint on the foot contacts requiring that their acceleration be negatively proportional to the velocity, and the constraints (\ref{con:contact_force},\ref{con:friction}) together ensure that contact forces remain within $\hat{K}$. The parameter vector $\vecg \eta$ allows bounded violations of the no-slip constraint to reduce the likelihood of infeasibility, $\varepsilon$ is a regularization constant typically set to a small value, e.g., $\varepsilon=10^{-8}$, and $\vec J = \partial \vec c / \partial \vec q$ is the Jacobian matrix for the vector of all contact points, $\vec c = [\begin{array}{ccc} \vec c_1^T &\dots& \vec c_{Nc}^T \end{array}]^T$. 


The weight parameter, $w_{\ddot{\vec q}}$, is used to balance the relative contribution of the desired motion cost with the ZMP tracking cost. To respect joint limits, the bounds $\ddot{q}_i \ge 0$ and $\ddot{q}_i \le 0$ are added for all $i$ such that $q_i = q_i^{\rm MIN}$ and $q_i = q_i^{\rm MAX}$, respectively. 

\section{Optimization}

We solve QP \ref{qp:controller} at each control step using a simple active-set method. The method assumes the set of active inequality constraints remains constant for consecutive solutions. It then produces a candidate solution by solving a partial set of optimality conditions derived from the assumed active set. If the candidate solution  satisfies the full set of optimality conditions, the assumption is correct and the algorithm terminates. Otherwise, the method updates the active set and repeats until a solution is found or a maximum number of iterations is reached. 

On rare occasions when no solution is found, the algorithm fails over to a more reliable (but on average slower) interior point solver. In our experiments, this lead to infrequent single-step input delays on the order of 3ms, which had no significant effect on walking performance. This contingency is required since finite termination cannot be guaranteed for the proposed method. In practice, however, instances of QP \ref{qp:controller} are almost always solved in one iteration. The computational cost of each iteration is also very small. A candidate solution is produced by solving a structured system of linear equations and constraints are evaluated only once. 

\subsection{Active-set method}
The QP solved at each control step can be written in the standard form,
\begin{align}
\begin{array}{cl}
 \stack{min}{\vec z} & \frac{1}{2} {\vec z}^T \vec W {\vec z}  +\vec g^T {\vec z}  \\ \mbox{subject to} & \vec A \vec z = \vec b  \\ & \vec P \vec z \le \vec f,
\end{array} \label{qp:template}
\end{align}
where the inequalities are defined by $\vec P=(\vec p_1, \vec p_2,\ldots,\vec p_n)^T$ and $\vec f = (f_1, f_2,\ldots, f_n)^T$. To solve this problem, it is assumed that $\vec p^T_i \vec z = f_i$ at optimality for each $i$ in a subset $\mathcal{A} \subseteq \{ 1 \ldots n \}$ called the \emph{active set}. For $t>0$, this subset equals the indices of the active inequalities from time $t-1$. With this assumption, the KKT conditions for the QP can be written in terms of $\vec z$, $\vecg \gamma$, and $\vecg \alpha$:
\begin{align}
\begin{array}{rclc}
 \vec W \vec z + \vec A^T \vecg \alpha + \sum_{i \in \mathcal{A}} \gamma_i 
 \vec p_i  &=& -\vec g  \\
\vec A \vec z &=& \vec b   \\
\vec p_i^T \vec z  &=& f_i & \forall i \in \mathcal{A} \\
 \gamma_i &=& 0 & \forall i \ne \mathcal{A}
\end{array} \label{sys:ls} \\
\begin{array}{rclc} 
 \vec P \vec z &\le& \vec f \\
 \gamma_i &\ge& 0 ~~& \forall i \in \mathcal{A}. \\
\end{array} \label{eq:optimality}
\end{align}

Our method solves the linear equations \eqref{sys:ls} and checks if the solution $(\vec z, \vecg \gamma,\vecg \alpha)$ satisfies the inequalities \eqref{eq:optimality}. If the inequalities are satisfied, $\vec z$ solves the  QP and the algorithm terminates.  
 Otherwise, the algorithm adds index $i$ to $\mathcal{A}$ if $\vec p_i^T \vec z > f_i$ or removes index $i$ if $\gamma_i < 0$ and resolves \eqref{sys:ls}.  The algorithm repeats this process until the inequalities \eqref{eq:optimality} are satisfied or a until a specified maximum number of iterations is reached. The method is outline in Algorithm~\ref{algo}.
 
\begin{algorithm}
 \SetAlgoLined
 \LinesNumbered
 \DontPrintSemicolon
 \KwData{A QP of form \eqref{qp:template} where the cost matrix $\vec W$ has the structure \eqref{eq:qStruct}. A set of constraints $\mathcal{A}$ assumed to be active at optimality.}
 \KwResult{An optimal solution $\vec z$ with active set $\mathcal{A}$ or a flag indicating failure.}
   $iter \leftarrow  0$ \\
 \Repeat{ $\vec z$ \rm and $\vecg \gamma$ \rm satisfy \eqref{eq:optimality}  }{
    Compute candidate solution $\vec z, \vecg \gamma, \vecg \alpha$ from (\ref{eq:linEq1},\ref{eq:linEq2}) \\ 
  \If {$\vec p^T_i \vec z > f_i$} {add $i$ to $\mathcal{A}$}
    \If {$\gamma_i < 0$} {remove $i$ from $\mathcal{A}$} 
    $iter \leftarrow iter + 1$ \\
    \If {$iter > iter_{\rm MAX}$} { \Return Failure } }
 \Return $\mathcal{A}$ and $\vec z$ \\ 
 \caption{Active-set method for solving \eqref{qp:template}. The set  $\mathcal{A}$ passed to the algorithm at time $t$ equals
the set of constraints active at optimality for time $t-1$. } 
\label{algo}
\end{algorithm}

\subsection{Efficiently computing a candidate solution } \label{subsec:findCand} 
The structure of QP~\ref{qp:controller} admits an efficient solution of the linear system \eqref{sys:ls}.  In particular, one can cheaply compute $\vec W^{-1}$ and construct a smaller system for $\vecg \alpha$ and $\vecg \gamma$.  Using a solution to this smaller system, one can then easily recover $\vec z$. To see this, first let $\vec P_{act}$ and $\vec f_{act}$ denote the rows of $\vec P$ and $\vec f$ indexed by $\mathcal{A}$ and let $\vec R= \begin{array}{cc} [\vec A^T & \vec P_{act}^T]^T \end{array}$ and $\vec e = [ \begin{array}{cc} \vec b^T & \vec f^T_{act} \end{array} ]^T$. A solution to \eqref{sys:ls} can be found by first solving the following system of equations for $\vecg \alpha$ and $\vecg \gamma$:
\begin{align}
 -\vec R \vec W^{-1} \vec R^T \left[ \begin{array}{c} \vecg \alpha \\ \vecg \gamma \end{array} \right] &= \vec e+\vec R \vec W^{-1} \vec g \label{eq:linEq1}  
\end{align}
Using a solution to this system, $\vec z$ can be recovered via
\begin{align}
 \vec z &= -\vec W^{-1} \left (\vec g + \vec R^T \left[ \begin{array}{c} \vecg \alpha \\ \vecg \gamma \end{array} \right]\right).  \label{eq:linEq2}
\end{align} 
Efficient computation of $\vec W^{-1}$ arises from its block diagonal structure,
\begin{align}
\vec W = \left[ \begin{array}{cc}  \vec W_{11}  &0 \\ 0 & \vec W_{22} \end{array} \right], \label{eq:qStruct} 
\end{align} 
where $\vec W_{22}$ is diagonal and $\vec W_{11} =  w_{\ddot{\vec q}} \vec I +  \vec U^T \vec Q \vec U$. For the ZMP dynamics, $\vec U=\vec D(t) \vec J \in \mathbb{R}^{2\times n}$, where $\vec J$ is the COM$(x,y)$ Jacobian and $\vec D(t)$ is the input mapping defined in (\ref{eq:tvzmp}). Applying the matrix inversion lemma yields an expression for $\vec W_{11}^{-1}$ that involves computing the inverse of $2\times2$ matrices:
\begin{align*}
\vec W_{11}^{-1}= \frac{1}{w_{\ddot{\vec q}}} \vec I - \frac{1}{w_{\ddot{\vec q}}^2}  \vec U^T(\vec Q^{-1}+ \frac{1}{w_{\ddot{\vec q}}} \vec U \vec U^T)^{-1} \vec U.
\end{align*}
It should also be noted that $\vec W^{-1}$ is independent of $\mathcal{A}$ and thus only needs to be computed once per control step even if multiple solver iterations are required. The same holds for various sub-matrices in the expressions \eqref{eq:linEq1} and \eqref{eq:linEq2}. 


\section{Application}

\begin{figure*}[thpb]
  \centering
  \includegraphics[scale=0.95]{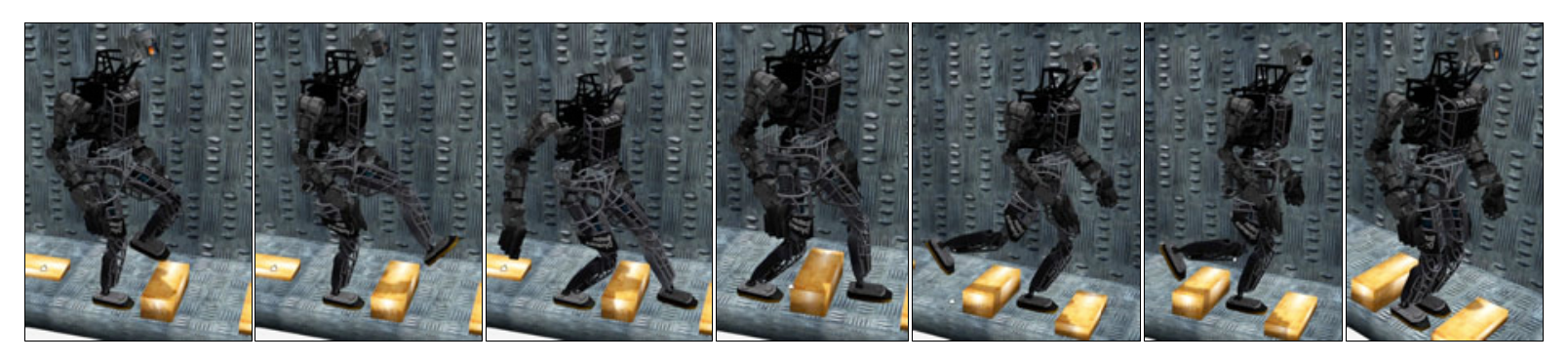}
  \includegraphics[scale=1.12]{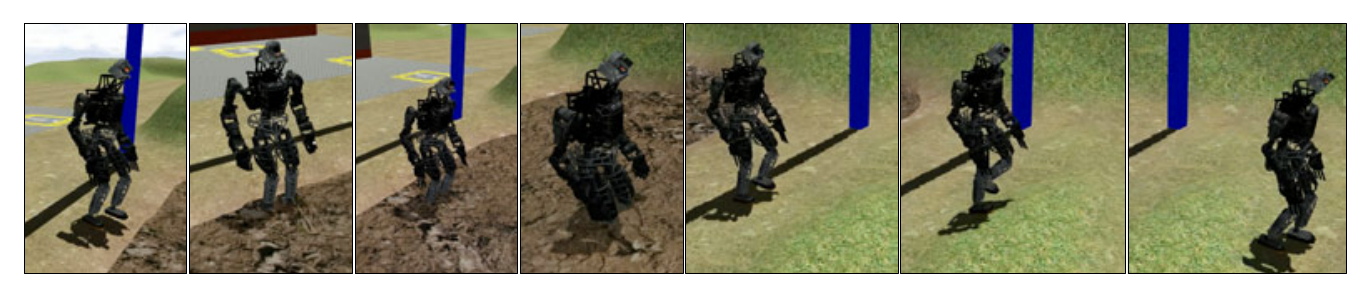}
  \caption{Walking in simulation over obstacles, through simulated mud, and over rolling hills using state and terrain estimation.}
  \label{fig:gazebo_walking}
 \end{figure*}

We implemented our controller using the 34-DOF Atlas humanoid model developed for the DARPA Virtual Robotics Challenge. Our evaluation of the controller included a variety of balancing and locomotion tasks using two independent simulation environments: Drake~\cite{Drake} and Gazebo~\cite{GazeboSim}. As part of MIT's entry into the DARPA Virtual Robotics Challenge (VRC), the controller was used to walk reliably over uneven terrain, through simulated knee-deep mud, and while carrying an unmodeled multi-link hose, all using imperfect state and terrain estimation (Figure~\ref{fig:gazebo_walking}).\footnote{Example simulation code is available at \\ \url{http://people.csail.mit.edu/scottk}.}

To design the balancing controller, we solved an infinite horizon LQR problem to regulate the ZMP at $(0,0)$. The cost functional took the form
\begin{eqnarray*}
J &=& \int_0^\infty \vec y^T \vec Q \vec y dt, \\
  &=& \int_0^\infty \left [\vec x^T \vec C^T \vec C \vec x + \vec u^T \vec D^T \vec D \vec u + 2\vec x^T \vec C^T \vec D \vec u \right]  dt,
\end{eqnarray*}
where $\vec Q = \vec I$. We assumed the COM height was constant while standing, thus making the ZMP dynamics linear. This had the advantage that it only required us to solve the LQR problem once. To see this, note that $J^*(\bar{\vec x})=\bar{\vec x}^T \vec S \bar{\vec x}$, where $\vec S$ is the solution of the algebraic Riccati equation. Thus the QP cost had the form,
\begin{eqnarray*}
\bar{\vec y}^T \bar{\vec y}  + 2\bar{\vec x}^T \vec S (\vec A \bar{\vec x} + \vec B \vec u) + w_{\ddot{\vec q}}|| \ddot{\vec q}_{\rm des}-\ddot{\vec q} ||^2 + \varepsilon \sum_{ij} \beta^2_{ij},
\end{eqnarray*}
where new desired ZMP locations $\vec k=[\begin{array}{cc} k_x &k_y \end{array}]^T$ could be achieved by a change in coordinates, $\bar{\vec y}=\vec y-\vec k$, $\bar{\vec x}=\vec x-\vec k$, and $\vec k$ is, e.g., the point at the center of the foot support polygon. In practice, we found the constant COM height assumption has minimal practical effect on balancing performance, even when recovery motions included significant hip bends and arm motion. We computed $\ddot{\vec q}_{\rm des}$ via a simple PD control rule, $\ddot{\vec q}_{\rm des} = K_p({\vec q}_{\rm des}- \vec q) - K_d(\dot{\vec q})$, using either a fixed nominal posture, ${\vec q}_{\rm des}$, for standing or a time-varying configuration trajectory for manipulation. We used the same scalar gains, $K_p$ and $K_d$, for all joints.


Our planning implementation took desired foot trajectories as input and computed a ZMP plan, $\vec y^d(t)$, by linear interpolation between step locations. The footstep planner combined terrain map information with heuristics to select reasonable step locations and timing. We solved the TVLQR problem (\ref{tvlqr_cost}) for the linear ZMP dynamics using the Riccati solution for balancing as the final cost, $\vec Q_f = \vec S$. The corresponding COM$(x,y)$ trajectory, $\vec x^d(t)$, can be computed by simulating the COM dynamics \eqref{com_dynamics} in a closed loop from time $t=0$ to $t=t_f$ with the optimal controller, $\bar{\vec u}^* = - \vec K(t) \bar{\vec x}$. In practice, we were able to compute both $J^*(\bar{\vec x},t)$ and $\vec x^*(t)$ for a 10m walking plan in approximately $1/4$s using an unoptimized MATLAB implementation of the explicit ZMP Riccati solution described by Tedrake et al. \cite{Tedrake_etalTBD}. 

The desired configuration, ${\vec q}_{\rm des}(t)$, was computed via inverse kinematics with constraints on the foot pose and COM position. Computation of ${\vec q}_{\rm des}(t)$ was done either offline for open-loop trajectory following or reactively inside the control loop by linearizing the forward kinematics at the current configuration and solving a second small QP to minimize the weighted $\ell^2$ distance to a nominal configuration while respecting foot pose, COM, and joint limit constraints. Qualitatively different motions could be achieved by varying the relative weights assigned to joints in the cost. For example, a smaller cost on back joints would tend to produce more torso sway to track the desired COM trajectory. 

We used a simplified 4-point contact representation for each foot. Active contacts were determined by a combination of the desired footstep plan and the estimated distance between the foot and terrain. If and only if the foot is perceived to be in contact and the plan agreed did we include the corresponding foot contact in the optimization. The requirement that both conditions be true was essential for breaking contact with the ground while walking. As with balancing, footstep and ZMP plans could be translated in three dimensions without additional computation by a simple change in coordinates in the QP cost.





\subsection{Solver Performance}

We compared the solve time of our active-set algorithm against two general-purpose QP solvers, Gurobi \cite{Gurobi} and CVXGEN \cite{CVXGEN}. For the Gurobi solver, we used the barrier (B) algorithm and dual simplex (DS) algorithm with both active constraint and solution warm-starting. Our CVXGEN problem formulation omitted the input saturation inequalities (\ref{con:inputs}) to fit within the problem size requirements. These experiments were done on an i7 2.1GHz quad-core laptop. A comparison of average solve times while executing a fixed flat ground pattern is given in Table~\ref{table:solve}.

\begin{table}[h]
\caption{Comparison of average QP solve times while walking. }
\label{table:solve}
\begin{center}
\begin{tabular}{|c|c|c|c|c|}
\hline 
  & Algorithm 1 & Gurobi (DS) & CVXGEN & Gurobi (B) \\ 
\hline 
Solve time & 0.2 ms & 1.0 ms & 2.2 ms & 3.1 ms \\ 
\hline 
\end{tabular}
\end{center}
\end{table}

\begin{figure}
  \centering
  \includegraphics[scale=0.4]{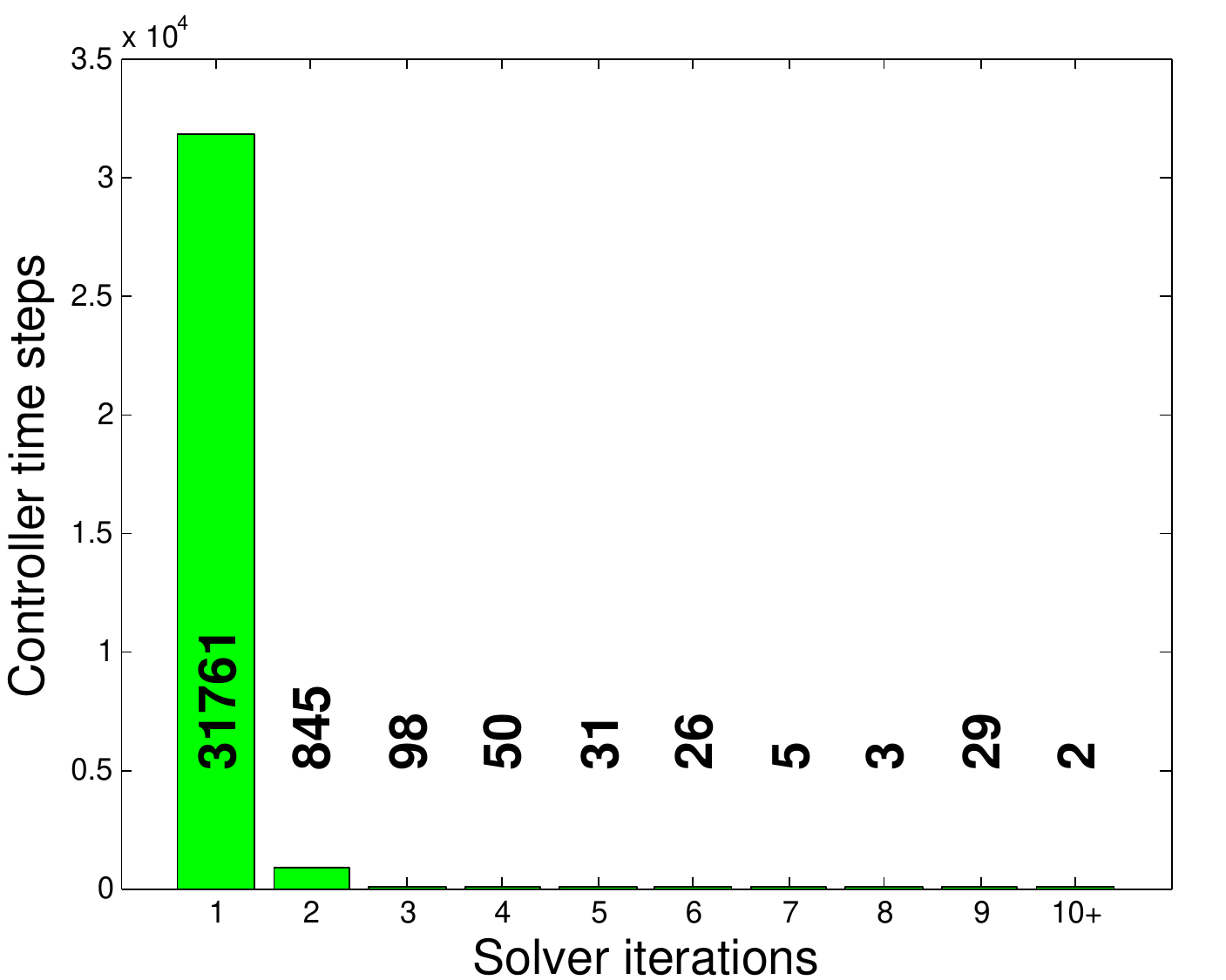}
\caption{Histogram of iterations needed to solve Quadratic Program \ref{qp:controller}  during a walking task. The method requires only one iteration approximately 97\% of the time.} \label{fig:hist}
\end{figure}

The custom active-set method outperforms the next best solver by a factor of 5. The significant performance advantage of Algorithm~\ref{algo} can be understood by considering the histogram in Figure~\ref{fig:hist}. For an overwhelming percentage of control steps, the active set does not change and the solver succeeds in a single iteration. Thus, most of the time control inputs are computed by solving a single linear system of equations. 

For the active-set algorithm, the total controller computation time is largely spent setting up the QP, which involves computing the manipulator dynamics, contact surface normals, and kinematic quantities such as the COM and contact Jacobians. In our implementation, the average QP setup time is approximately $0.8$ms for the 34-DOF Atlas model, giving us a total control step time of $1$ms. 

\begin{figure}
  \centering
  \includegraphics[scale=0.65]{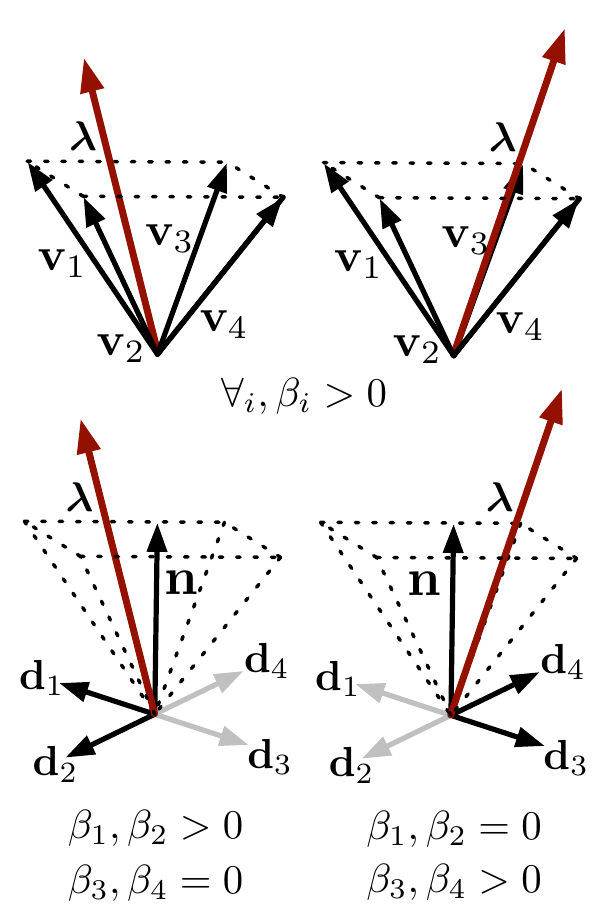}
\caption{An illustration showing how different approximate friction cone parameterizations can affect active set stability.} \label{fig:friction}
\end{figure}

The performance of the solver does have a subtle dependency on the problem formulation. We found that using the parameterization of the approximate friction cone (\ref{k_approx}) lead to fewer active set changes than the commonly used Stewart and Trinkle~\cite{Stewart96} parameterization,
\begin{eqnarray}
\hat{K}_{\rm ST} = \left \{ z \vec n + \sum_{i=1}^{N_d} \beta_i \vec d_i: z \ge 0, \beta_i \ge 0, \sum_{i=1}^{N_d} \beta_i \le \mu z \right \},\label{k_stewart}
\end{eqnarray}
where we have dropped the explicit contact point index, $j$. The parameterization \eqref{k_stewart} lead to approximately 50\% more control steps requiring 2 iterations or more. Intuitively, this is a result of the fact that the active inequalities constraints, $\{i : \beta_i = 0\}$, under parameterization \eqref{k_stewart} can change when forces inside the approximate friction cone change direction. By contrast, when using \eqref{k_approx}, the constraints on $\beta_i$ only become active on the surface of the polyhedron. This idea is illustrated in Figure~\ref{fig:friction}.

\section{Related Work}

The controller design we proposed shares some properties with other horizon-1 MPC implementations. For example, the same flavor of dynamic, friction, and foot motion constraints have appeared in other QP formulations \cite{Abe07, Herzog_etal2013,Saab_etal2013, Koolen_etal2013}. 
Herzog et al. \cite{Herzog_etal2013} proposed the idea of separating the manipulator equation into floating-base and actuated DOFs to remove $\vecg \tau$ as a decision variable, which enabled them to achieve control rates of 1kHz for a 14-DOF biped. Polyhedral approximations are frequently used to linearize friction constraints, but to our knowledge no prior connection has been made between different parameterizations and solver performance. 

Ames et al. \cite{Ames_etal2012,AmesHSCC2013} used CLFs for walking control design by solving QPs that minimize the input norm, $||\vec u|| $,  while satisfying constraints on the negativity of $\dot{V}_{\rm clf}$. By contrast, we placed no constraint on $\dot{V}_{\rm clf}$ and instead minimized an objective of the form $\ell(\vec x, \vec u) + \dot{V}_{\rm clf}$, where $\ell(\vec x, \vec u)$ is an instantaneous cost on $\vec x$ and $\vec u$. This approach gave us the significant practical robustness while making the QP less prone to infeasibilities. 

Other uses of active-set methods for MPC have exploited the temporal relationship between the QPs arising in MPC. Bartlett et al. compared active-set and interior-point strategies for MPC \cite{Bartlett_etal2000}. The described an active-set approach based on Schur complements for efficiently resolving KKT conditions after changes are made to the active set. This framework is analogous to the solution method we discuss in Section \ref{subsec:findCand}. In the discrete time setting, Wang and Boyd ~\cite{WangBoyd2011} describe an approach to quickly evaluating control-Lyapunov policies using explicit enumeration of active sets in cases where the number of states is roughly equal to the square of the number of inputs. 

Ferreau et al. \cite{Ferreau2008} consider the MPC problems where the cost function and dynamic constraints are the same at each time step; i.e., the QPs solved at iteration differ only by a single constraint that enforces initial conditions. By smoothly varying the initial conditions from the previous to the current state, they were able to track a piecewise linear path traced by the optimal solution, where knot points in the path correspond to changes in the active set. Since the controller we considered had changing cost and constraint structure, this method would have been difficult to apply.

\section{Conclusion}

We described a stabilizing QP controller formulation for dynamic walking and solution technique that exploits consistency between active inequality constraints in subsequent control steps. In our experiments with a simulated Atlas robot, we were able to efficiently compute control inputs while walking by solving a single system of linear equations a high-percentage of the time, hence outperforming several popular general-purpose solvers used frequently in the literature. Although we have focused on humanoids and ZMP dynamics in this paper, the QP formulation we described is equally applicable to more general floating-base systems and other types of simple system models. 
Similarly, the active-set method used in this work could easily be applied to the various MPC formulations that exist in the literature. Our current efforts are focused on adapting this approach to achieve stable walking, climbing, and manipulation with a physical Atlas humanoid robot at MIT.  

\addtolength{\textheight}{-12cm}   




\section*{Acknowledgments}

We would like to thank the members of the MIT VRC team for their contributions to the perception and estimation algorithms that made walking in the simulation challenge possible. We thank Robin Deits for designing the footstep planner used by the controller described in this paper. 


\bibliographystyle{IEEEtran}
\bibliography{myrefs}		

\end{document}